\documentclass[runningheads]{llncs}
\usepackage{amssymb}
\usepackage{amsmath}
\usepackage{graphicx}
\usepackage{subcaption}
\usepackage{hyperref} 
\usepackage{float}
\usepackage{dsfont}
\usepackage{calrsfs}
\usepackage{xcolor}
\usepackage{graphicx}
\usepackage{tikz}
\usepackage{wrapfig}
\usepackage[T1]{fontenc} 

\usepackage{array}
\usepackage{pifont}
\DeclareMathAlphabet{\pazocal}{OMS}{zplm}{m}{n}

\usepackage{multirow}
\usepackage{amsmath}
\usepackage{caption}
\newcolumntype{P}[1]{>{\centering\arraybackslash}p{#1}}

\makeatletter
\newcommand\footnoteref[1]{\protected@xdef\@thefnmark{\ref{#1}}\@footnotemark}
\makeatother

\begin{document}
\title{Debiasing Counterfactuals In the Presence of Spurious Correlations}

\titlerunning{Counterfactual Debiasing}

\author{Amar Kumar$^{1,2}$, Nima Fathi$^{1,2}$, Raghav Mehta$^{1,2}$, Brennan Nichyporuk$^{1,2}$, Jean-Pierre R. Falet$^{2,3}$, Sotirios Tsaftaris $^{4,5}$, Tal Arbel$^{1,2}$}
\authorrunning{A. Kumar et al.}
\institute{$^1$Center for Intelligent Machines, McGill University, Canada. \\ $^2$MILA (Quebec AI institute), Canada. \\ $^3$Montreal Neurological Institute, McGill University, Canada. \\ $^4$Institute for Digital Communications, School of Engineering, University of Edinburgh, UK. \\ $^5$The Alan Turing Institute, UK\\
\email{amarkr@cim.mcgill.ca}}

%
%
\maketitle              
\begin{abstract}
Deep learning models can perform well in complex medical imaging classification tasks, even when basing their conclusions on spurious correlations (i.e. confounders), should they be prevalent in the training dataset, rather than on the causal image markers of interest. This would thereby limit their ability to generalize across the population. Explainability based on counterfactual image generation can be used to expose the confounders but does not provide a strategy to mitigate the bias. In this work, we introduce the first end-to-end training framework that integrates both (i) popular debiasing classifiers (e.g. distributionally robust optimization (DRO)) to avoid latching onto the spurious correlations and (ii) counterfactual image generation to unveil generalizable imaging markers of relevance to the task. Additionally, we propose a novel metric, {\it Spurious Correlation Latching Score (SCLS)}, to quantify the extent of the classifier reliance on the spurious correlation as exposed by the counterfactual images.  Through comprehensive experiments on two public datasets (with the simulated and real visual artifacts), we demonstrate that the debiasing method: (i) learns generalizable markers across the population, and (ii) successfully ignores spurious correlations and focuses on the underlying disease pathology.
\end{abstract}
\section{Introduction}
Deep learning models have shown tremendous success in disease classification
based on medical images, given their ability to learn complex imaging markers across a wide population of subjects. 
These models can show good performance and still be {\it biased} as they may focus on spurious correlations in the image that are not causally related to the disease but arise due to confounding factors - should they be common across the majority of samples in the training dataset. As a result,  the confounding predictive image markers may not generalize across the population. For example, a deep learning model was able to accurately detect COVID-19 from chest radiographs, but rather than relying on pathological evidence, the model latched on to spurious correlations such as medical devices or lettering in the image~\cite{degrave2021ai}. As a result, these image markers did not generalize across the population.

In order to safely deploy black-box deep learning models in real clinical applications, explainability should be integrated into the framework so as to expose the spurious correlations  on which the classifier based its conclusions. Popular post-hoc explainability strategies, such as Grad-CAM~\cite{panwar2020deep,jiang2020multi,selvaraju2017grad}, SHAP~\cite{lundberg2017unified}, LIME~\cite{magesh2020explainable} are not designed to expose the precise predictive markers driving a classifier. Models that integrate counterfactual image generation, along with black-box classifers~\cite{singla2023explaining,cohen2021gifsplanation,thiagarajan2022training}, permit exposing the predictive markers used by the classifier. However, should 
these methods discover that the markers are indeed simply visual artifacts there are no strategies to mitigate the resulting biases. Furthermore, although several debiasing methods have been successfully implemented to account for generalizability~\cite{burlina2021addressing,zong2022medfair,ricci2022addressing,larrazabal2020gender,zou2018ai}, they do not integrate explainability into the framework in order to provide reasons for improved performance. 

Therefore, the important question to be answered is - \textit{Can a model be trained to disregard spurious correlations and identify generalizable predictive disease markers?}

\begin{figure}[h]
    \vspace{-5mm}
    \centering
    \includegraphics[width=0.8\textwidth]{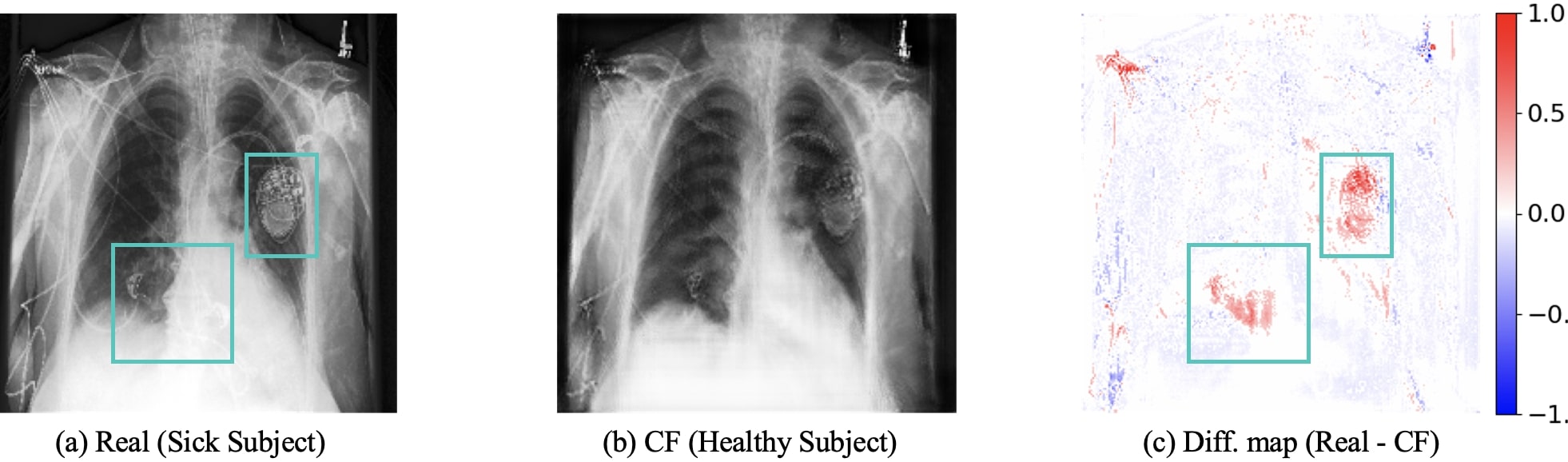}
    \scriptsize
    \caption{Counterfactual (CF) image indicating that the classifier latched onto spurious correlations (medical devices) when correctly predicting that subject is sick (class: Pleural Effusion), due to their prevalence in the training dataset for this class. (a) Chest radiograph of a sick subject with several medical devices shown (cyan boxes), (b) Generated (CF) image,  
    (c) Difference heat map shows maximum change around the medical devices, rather than indicating the correct markers for the disease. 
    }
    \label{fig:intro_problem}
    \vspace{-5mm}
\end{figure}

In this paper, we propose the first end-to-end training framework for the explainability of classifier and debiasing via counterfactual image generation. We seek to discover imaging markers that reflect underlying disease pathology and that generalize across subgroups. Extensive experiments are performed on two different publicly available datasets - (i) \textit{\href{https://www.kaggle.com/c/rsna-pneumonia-detection-challenge/}{RSNA Pneumonia Detection Challenge}} and (ii) \textit{CheXpert} \cite{irvin2019chexpert}. To illustrate the goal, Figure~\ref{fig:intro_problem} shows an example from the contrived CheXpert dataset, where most of the sick subjects have medical device(s) (e.g. a pacemaker) in their images while most of the healthy subjects do not. As such, there exists a spurious correlation between a confounding visual artifact (the medical devices) and the disease. A classifier based on a standard optimization technique, empirical risk minimization (ERM), incorrectly indicates the medical device as a disease marker, as depicted by the counterfactual (CF). In this work, we propose replacing ERM with a popular debiasing method, Group-DRO (distributional robust optimization). This permits the classifier to focus on the pathological image markers of the disease rather than on spurious correlation(s). Additionally, we show that Group-DRO ignores the visual artifact when making its decision, and  generalizes across subgroups without the spurious correlation. Since standard metrics to evaluate counterfactuals do not indicate the region where the classifier focuses, we also propose a novel metric, the Spurious Correlation Latching Score (SCLS), to measure the degree to which the classifier latches onto spurious correlations. Our experiments indicate an improvement (in terms of differences in classifier outputs) of 0.68 and 0.54 in the SCLS using the Group-DRO classifier over the ERM 
for each of the two datasets.

\section{Methodology}
\label{sec:Method}
We propose an end-to-end training strategy to explain the output of a classifier. Here, we are considering a scenario where  
majority of the training data encompasses a spurious correlation with the target label. However, there is also a minority subgroup in the dataset that does not have any spurious correlation with the target label i.e., if the classifier was to rely onto the spurious correlation then the performance on these minority subgroups will be poor. Also, the term `majority' and `minority' is based on the number of samples in these groups. An overview of our approach is shown in Figure \ref{fig:overview}. 

\subsection{Classifier Explainability and Debiasing Via Counterfactual Image Generation}
\noindent\textbf{\textit{Disease Classification}}
Binary (e.g. "sick" or "healthy") classification of the images is performed using either a standard classifier (ERM~\cite{vapnik1991principles}), or a classifier that mitigates biases across sub-groups (Group-DRO~\cite{sagawa2019distributionally}).  The ERM classifier ($f_{ERM}$) is expected to be affected by the spurious correlation present in the training dataset, as it minimizes the loss over the entire training dataset and latching onto spurious correlation is a shortcut to minimize the loss. Thus, it would not generalize across the minority subgroups of the dataset~\cite{mehta2023evaluating,shui2023mitigating}. On the contrary, the DRO classifier ($f_{DRO}$) is not expected to learn the spurious correlation as it considers the majority and minority subgroups separately when optimizing the loss. Thus, it would generalize well across all subgroups. \\

\begin{figure}[t]
    \centering
     \scriptsize
    \includegraphics[scale=0.25]{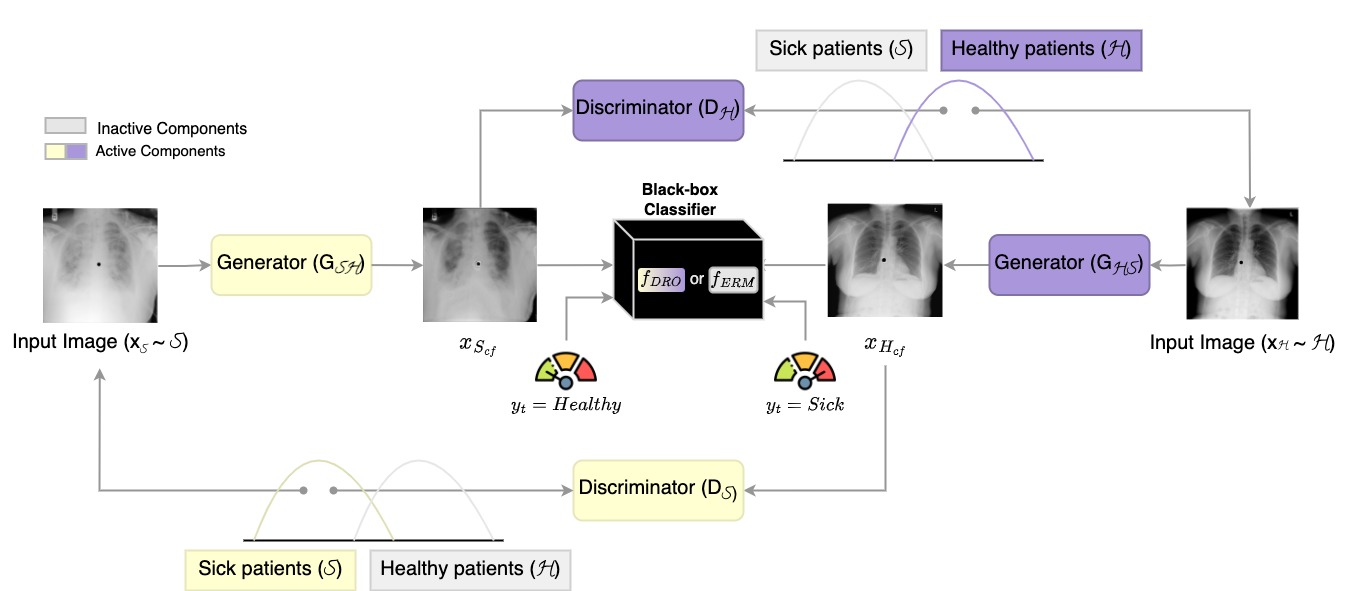}
    \scriptsize
    \caption{Training procedure overview: The black-box classifier can be $f_{ERM}$ or $f_{DRO}$ and provides supervision to maintain the correct target class, $y_t$. Two U-Net generators, $G_{\pazocal{S}\pazocal{H}}$ and $G_{\pazocal{H}\pazocal{S}}$, are employed to synthesize counterfactual images, namely $x_{\pazocal{S}_{cf}}$ and $x_{\pazocal{H}_{cf}}$. The discriminator $D_{\pazocal{H}}$ and $D_{\pazocal{S}}$ compares the counterfactual images with the domain of healthy $\pazocal{H}$ and sick $\pazocal{S}$ subjects respectively. Note, training a cycle-GAN requires simultaneous use of two input images from the two distributions.}
    \label{fig:overview}
    \vspace{-5mm}
\end{figure}

\noindent\textbf{\textit{Generative model for synthesizing counterfactuals}}
We develop an explainability framework that integrates counterfactual image generation together with a classifier during training. We adapted Cycle-GAN~\cite{zhu2017unpaired} as the generative model for counterfactual image generation, chosen for its strong performance across a variety of domains~\cite{mertes2022ganterfactual,zhu2017unpaired}. A pre-trained, frozen binary classifier ($f_{ERM}$ or $f_{DRO}$) provides supervision to the generator. The proposed architecture and optimization objectives (see Figure~\ref{fig:overview}) are designed to generate counterfactual images that adhere to the following common constraints~\cite{mothilal2020explaining,nemirovsky2020countergan,kumar2022counterfactual}: (i)\textit{Identity preservation}: The counterfactual images resemble the input images with minimal change; (ii) \textit{Classifier consistency}: Counterfactual images belong to the target class; (iii)\textit{Cycle consistency}: When counterfactual images are fed through the opposing  generator, the output reverts to the original image (see Figure~\ref{fig:overview}).  

During inference, based on the classifier's decision (i.e., $f_{ERM} \text{ or } f_{DRO}$) for the input image, we generate counterfactual images and analyze the difference heatmap between the factual (input) and counterfactual (synthesized) images. This interpretable heat map indicates the image markers that contribute the most to changing the classifier's decision.

\subsection{Metrics for Evaluating Counterfactuals: Accounting for Spurious Correlations}\label{EvalCoun}

Standard counterfactual evaluation metrics are structured so as to ensure that the generated images (a) preserve the subject identity and thus penalize generated counterfactual images that are significantly different from the factual (original) images and (b) result in a maximal change in the class label (e.g. from healthy to sick). Identity preservation is typically measured by \textit{structural similarity index } (SSIM)~\cite{hore2010image} and \textit{Actionability} ~\cite{nemirovsky2020countergan,mothilal2020explaining}, defined as $\mathbb{E}\left[\left \| x - x_{cf}  \right \|_{L_1}\right]$ between factual ($x$) and counterfactual ($x_{cf}$) images. Here, a higher value for SSIM and a lower value for Actionability would indicate better counterfactuals.
The \textit{counterfactual prediction gain} (CPG)~\cite{nemirovsky2020countergan}, defined as $|f(x)-f(x_{cf})|$, indicates the degree of change in the classifier's decision such that a higher value of CPG indicates better counterfactuals. 

While such metrics are required to measure the validity of the generated counterfactuals, they do not assess whether the classifier latched onto spurious correlations. For example, consider an image of a sick subject in the presence of a spurious correlation. If the disease classifier, $f_{ERM}$, latched onto the spurious correlation when identifying the subject as sick, the corresponding counterfactual image (i.e., depicting a healthy subject) would show changes in the area of the spurious correlation. In this case, all three evaluation metrics mentioned above would determine that this is a valid counterfactual image, based on high SSIM and low Actionability (shows minimal changes made compared to the factual image) and high CPG (due to the classifier decision changing from sick to healthy). However, the counterfactual image shows changes in the area of the spurious correlation rather than depicting the correct predictive image markers for the disease as desired. 

In order to indicate that the classifier is correct but for the wrong reasons, we introduce a novel metric called Spurious Correlation Latching Score (SCLS) defined as follows: 
\begin{equation}
    \text{SCLS} = |d(x)-d(x_{cf})|. 
\end{equation} 
Here, $d(\cdot)$ is a separate classifier, trained to identify the presence of spurious correlation in the image. In cases where the counterfactual image makes changes in an area of spurious correlation, SCLS will be high, as the  $d(\cdot)$  will show a maximum change in its prediction between factual and counterfactual images. On the other hand, if the counterfactual image does not make changes in the area of the spurious correlation then SCLS will have a low value. As such, this evaluation strategy will validate how well the counterfactuals can help to determine that the classifier latched onto spurious correlations.

\section{Experiments and Results} \label{section:dataset_exp_results}

\subsection{Dataset and Implementation Details}

We perform experiments on two publicly available datasets.
The absence of ground truth makes the validation of counterfactual images particularly challenging. Therefore, to directly evaluate the quality of the generated counterfactual images in the presence of spurious correlations, we modify a publicly available dataset (\textit{\href{https://www.kaggle.com/c/rsna-pneumonia-detection-challenge/}{RSNA Pneumonia Detection Challenge}}) by adding a synthetic artifact to the majority of the sick images (90\%).  
The majority of the sick and few of the healthy subjects have an artifact in the image, whereas the majority of the healthy and a few sick subjects do not have this artifact. The spurious correlation (artifact) is a black dot of radius 9 pixels at the center of the image.
Thus, there are a total of four subgroups ($majority_S$, $majority_H$, $minority_S$ and $minority_H$)
in the dataset with varying number of images: $majority_S$ and $majority_H$ are majority subgroups (sick with artifact and healthy without artifact), while $minority_S$ and $minority_H$ are minority subgroups (sick without artifact and healthy with artifact). Henceforth, this dataset will be referred to as Dataset 1.

\begin{table}[h]
\centering
\scriptsize
\begin{tabular}{ccccc}
\hline
& \textbf{Disease} & \textbf{Image size} & \textbf{Classifier}                       & \begin{tabular}[c]{@{}c@{}} \textbf{\# samples} \\ \textbf{{[}$majority_S$, $minority_S$,}\\ \textbf{$minority_H$, $majority_H${]}}\end{tabular} \\ \hline
\textbf{Dataset 1} & Pneumonia                                                   & 512 x 512  & AlexNet~\cite{mertes2022ganterfactual} & 5413, 1526, 883, 7968 \\ 
& & & &  \\ 
\textbf{Dataset 2} & \begin{tabular}[c]{@{}l@{}}Pleural \\ Effusion\end{tabular} & 224 x 224  & \begin{tabular}[c]{@{}l@{}}Resnet-50~\cite{targ2016resnet}\\ (pre-trained)\end{tabular} & 2600, 260, 350, 3456 \\ \hline
\end{tabular}
\scriptsize
\caption{Implementation details for the two datasets}
\label{tab:implementation}

\end{table}

\begin{figure}[t]
    \vspace{-2mm}
    \centering
    \includegraphics[scale=0.22]{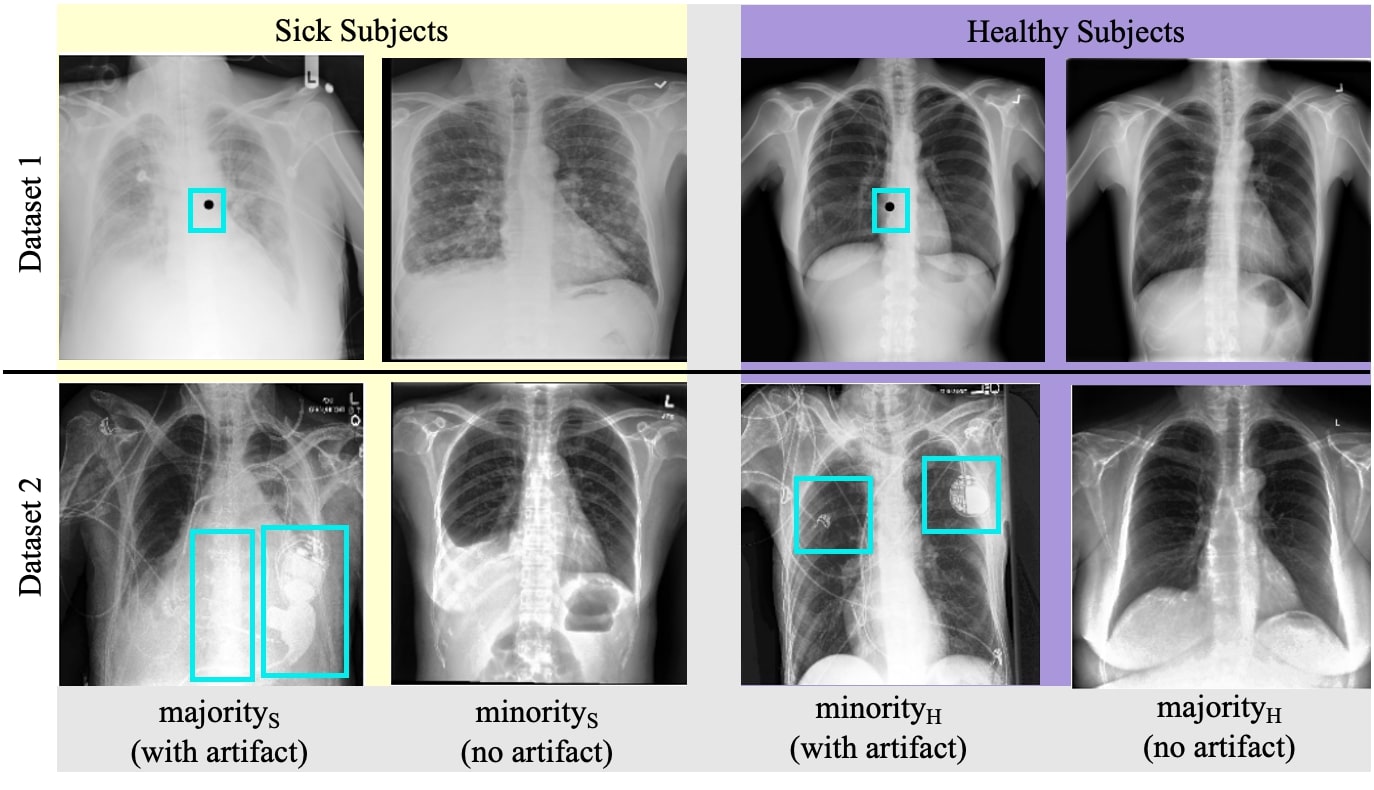}
    \caption{Datasets 1 and 2 group division: The majority of the sick subjects [$majority_S$] and the minority of healthy subjects [$minority_H$] have visual artifacts (shown in cyan boxes). The majority of healthy subjects [$majority_H$] and the minority of sick subjects [$minority_S$] do not have visual artifacts. Top row: Simulated artifacts (black dots); Bottom row: Real artifacts (medical devices). }
    \label{fig:subgroups}
\vspace{-5mm}
\end{figure}

We also show experiments on a subset of a publicly available dataset (\textit{CheXpert} \cite{irvin2019chexpert}) with medical devices (visual artifacts), spuriously correlated with the disease. Specifically, we extract the subset of images that have labels ``healthy'' or ``pleural effusion'' (subjects with the presence of other diseases are removed from the dataset). This dataset will be referred to as Dataset 2. More details about both datasets are provided in  Table~\ref{tab:implementation}. Note that both the datasets are divided into training/validation/testing with 70/10/20 random split.
Example images for both datasets and all four subgroups are shown in Figure~\ref{fig:subgroups}. 

\subsection{Results}
\begin{figure}[t]
    \centering
    \scriptsize
    \includegraphics[scale=0.17]{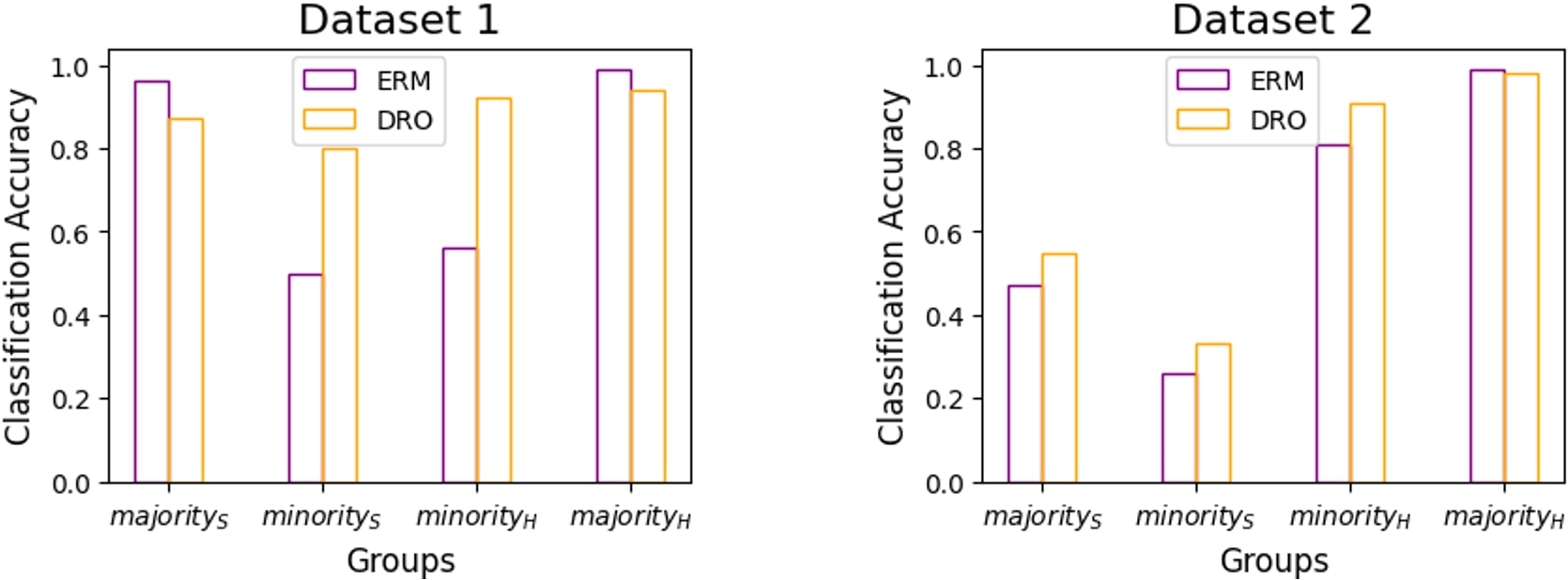}
    \caption{Performance of ERM ($f_{ERM}$) and DRO ($f_{DRO}$) based classifier on a held out test set across all subgroups for both datasets. Notice that DRO has improved performance on minority subgroup {[}$minority_S$ and $minority_H${]} showing improved generalizability across all subgroups.}
    \label{fig:erm_dro_prf}
    \vspace{-5mm}
\end{figure}

\noindent\textbf{Classifier Evaluation} For both datasets (Figure~\ref{fig:erm_dro_prf}), the DRO-based classifier ($f_{DRO}$) performs better for the minority subgroups ($minority_S$ and $minority_H$); indicating that it can better generalize to sub-populations that do not have the same visual artifact as the majority subgroups. Both classifiers perform similarly for the majority subgroups ($majority_S$ and $majority_H$).

\begin{figure}[h!]
    \centering
    \scriptsize
    \includegraphics[scale=0.25]{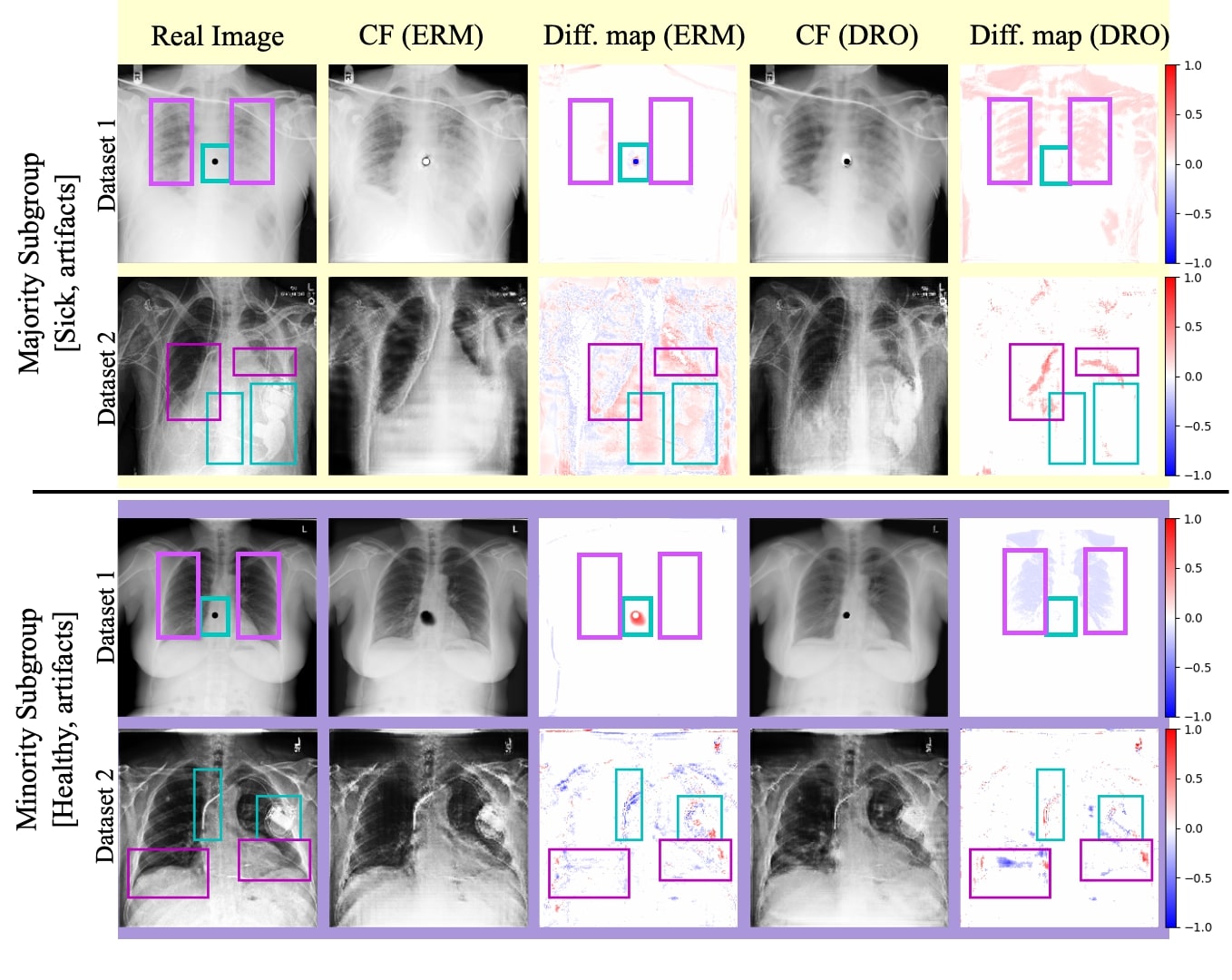}
     
    \caption{Qualitative comparison of counterfactual (CF) images generated with ERM and DRO classifiers for both majority (top row) and minority (bottom row) subgroups. The ERM-based CFs show significant changes in the areas of spurious correlation (cyan boxes), whereas the DRO-based CFs show almost no changes in the same areas. In contrast, significant changes can be seen in the expected area of disease pathology (magenta boxes) in DRO-based CFs, while the ERM-based CFs show little to no changes in these areas.} 
    \label{fig:qualitative}
    \vspace{-5mm}
\end{figure}

\noindent\textbf{{Qualitative Counterfactual Evaluation}} Pneumonia in chest radiograph manifests as increased brightness in some regions of the lungs. In dataset 1, when examining the majority subgroup of sick subjects, the ERM-based classifier latches onto the spurious correlation, as seen by the difference maps. On the other hand, a DRO-based classifier focuses on the pathology of the disease, indicated by darker intensity regions over the lungs, as shown in Figure~\ref{fig:qualitative}. The behavior of $f_{ERM}$ is also evident in the minority subgroup, where the counterfactual for a healthy subject exhibits an enlarged artifact, wrongly suggesting that the visual artifact serves as a disease marker. 
Pleural effusion is characterized by the rounding of the costophrenic angle, augmented lung opacity, and reduced clarity of the diaphragm and lung fissures~\cite{light2002pleural}. For the majority subgroup of sick subjects in Dataset 2, the counterfactual images based on ERM remove the medical device rather than focusing on the disease. In addition, for healthy subjects from the minority subgroup, maximum changes are observed around the medical device. On the other hand, for the majority subgroup, the DRO-based counterfactuals show changes around the expected areas while preserving the medical device.

\noindent\textbf{{Quantitative Counterfactual Evaluation}}
In Table~\ref{table:quant}, counterfactual images generated by ERM and DRO show similar scores according to standard metrics:  SSIM, Actionability and CPG. As these metrics are not designed to quantify whether the generated counterfactuals are affected by spurious correlations (see Section~\ref{EvalCoun}), the quality of the counterfactuals is now examined based on the proposed SCLS metric. The AUC of the classifier, $d$, trained to detect the presence of artifacts is 1.0 and 0.82 for Dataset 1 and Dataset 2, respectively. As indicated by the last row of Table~\ref{table:quant}, the ERM-based classifier shows a high value (poor performance) for SCLS for both datasets. On the other hand, the DRO-based classifier has a low value (good performance) for SCLS for both datasets. These results corroborate the finding made by visual comparison of the counterfactual images generated by the ERM and DRO classifiers. 
Overall, both qualitative and quantitative evaluations indicate that an ERM-optimized classifier latches on to the spurious correlation prevalent in the dataset, while a DRO-optimized classifier can be trained to successfully ignore the spurious correlation. 

\begin{table}
\centering
\setlength{\tabcolsep}{8pt}
\begin{tabular}{l|cc|cc}
\hline
    & \multicolumn{2}{c|}{\textbf{Dataset 1}}   & \multicolumn{2}{c}{\textbf{Dataset 2}}   \\ \cline{2-5} 
    & {ERM}  & DRO  & {ERM}  & DRO  \\ \hline

{Actionability $\downarrow$} & {7.68 $\pm$ 0.01}  & 7.86 $\pm$ 0.01  & {4.93 $\pm$ 0.01}  & 5.68 $\pm$ 0.04  \\ 
{SSIM $\uparrow$}   & {98.03 $\pm$ 0.00} & 98.44 $\pm$ 0.01 & {98.21 $\pm$ 0.01} & 98.36 $\pm$ 0.01 \\ 
{CPG $\uparrow$} & {0.91 $\pm$ 0.04} & 0.96 $\pm$0.03 & {0.88 $\pm$ 0.07} & 0.89 $\pm$ 0.04\\ \hline \hline
{\textbf{SCLS} $\downarrow$}   & {0.80 $\pm$ 0.08} &  $\mathbf{0.12 \pm 0.07}$ 
& {0.76 $\pm$ 0.09} & $\mathbf{0.22 \pm 0.06}$\\ \hline

\end{tabular}
\caption{Quantitative results to compare counterfactual images generated for both datasets. A low SCLS value implies that the model ($f_{DRO}$ in this case) did not latch onto the spurious correlation.}\label{table:quant}
\vspace{-10mm}
\end{table}

\section{Conclusion}
Safe deployment of black-box models requires explainability to disclose when the classifier is basing its predictions on spurious correlations and is therefore not generalizable. In this paper, we presented the first integrated end-to-end training strategy for generating unbiased counterfactual images, capitalizing on a DRO classifier to enhance generalization. Our experiments based on two datasets demonstrate that, unlike standard ERM classifiers which are susceptible to latching onto spurious correlations, the unbiased DRO classifier performs significantly better for minority subgroups in terms of- (a) the classifier performance and (b) the novel SCLS metric, which quantifies the degree to which the classifier latches on to the spurious correlation as depicted by the generated counterfactual images. 
 
Current datasets typically do not provide the ground truth predictive markers of interest. Future work will require localizing the predictive markers (e.g. with bounding boxes) and determining the degree of overlap with the discovered markers. 
Further, we intend to explore the power of alternative debiasing techniques and their potential contribution to discovering generalizable image markers. 

\subsubsection{Acknowledgements} The authors are grateful for funding provided by the Natural Sciences and Engineering Research Council of Canada, the Canadian Institute for Advanced Research (CIFAR) Artificial Intelligence Chairs program, the Mila - Quebec AI Institute technology transfer program, Microsoft Research, Calcul Quebec, and the Digital Research Alliance of Canada. S.A.\ Tsaftaris acknowledges the support of Canon Medical and the Royal Academy of Engineering and the Research Chairs and Senior Research Fellowships scheme (grant RCSRF1819 /\ 8 /\ 25), and the UK’s Engineering and Physical Sciences Research Council (EPSRC) support via grant EP/X017680/1.

\newpage
\bibliographystyle{splncs04}
\bibliography{paper24}

\end{document}